\documentclass[conference]{IEEEtran}

\usepackage{microtype}
\usepackage{graphicx}
\usepackage{subfigure}
\usepackage{booktabs} 

\usepackage{hyperref}

\usepackage{multirow,makecell}
\usepackage{amsmath}
\usepackage{algorithmic}
\usepackage{diagbox}

\usepackage{graphicx}
\usepackage{subfigure}
\usepackage{multirow}
\usepackage{xcolor}
\usepackage{booktabs}
\usepackage{amssymb}

\newcommand{\M}{VAQF}

\usepackage{amsmath}
\usepackage{amssymb}
\usepackage{mathtools}
\usepackage{amsthm}

\usepackage[capitalize,noabbrev]{cleveref}

\theoremstyle{plain}

\theoremstyle{definition}

\theoremstyle{remark}

\usepackage[textsize=tiny]{todonotes}

\begin{document}

\title{\M: Fully Automatic Software-Hardware Co-Design Framework for Low-Bit Vision Transformer}

\author{
Mengshu Sun\textsuperscript{\rm 1}, Haoyu Ma\textsuperscript{\rm 2}, Guoliang Kang\textsuperscript{\rm 3}, Yifan Jiang\textsuperscript{\rm 3}, Tianlong Chen\textsuperscript{\rm 3}, Xiaolong Ma\textsuperscript{\rm 1}, \\ Zhangyang Wang\textsuperscript{\rm 3}, Yanzhi Wang\textsuperscript{\rm 1} \\
\textsuperscript{\rm 1}Northeastern University
\textsuperscript{\rm 2}University of California, Irvine
\textsuperscript{\rm 3}University of Texas at Austin \\
\{sun.meng, ma.xiaol, yanz.wang\}@northeastern.edu haoyum3@uci.edu \\  kgl.prml@gmail.com  \{yifanjiang97, tianlong.chen, atlaswang\}@utexas.edu}

\maketitle

\begin{abstract}
The transformer architectures with attention mechanisms have obtained success in Nature Language Processing (NLP), and Vision Transformers (ViTs) have recently extended the application domains to various vision tasks. While achieving high performance, ViTs suffer from large model size and high computation complexity that hinders the deployment of them on edge devices. To achieve high throughput on hardware and preserve the model accuracy simultaneously, we propose \M, a framework that builds inference accelerators on FPGA platforms for quantized ViTs with binary weights and low-precision activations. Given the model structure and the desired frame rate, \M~will automatically output the required quantization precision for activations as well as the optimized parameter settings of the accelerator that fulfill the hardware requirements. The implementations are developed with Vivado High-Level Synthesis (HLS) on the Xilinx ZCU102 FPGA board, and the evaluation results with the DeiT-base model indicate that a frame rate requirement of 24 frames per second (FPS) is satisfied with 8-bit activation quantization, and a target of 30 FPS is met with 6-bit activation quantization.
To the best of our knowledge, this is the first time quantization has been incorporated into ViT acceleration on FPGAs with the help of a fully automatic framework to guide the quantization strategy on the software side and the accelerator implementations on the hardware side given the target frame rate. Very small compilation time cost is incurred compared with quantization training, and the generated accelerators show the capability of achieving real-time execution for state-of-the-art ViT models on FPGAs.
\end{abstract}

\section{Introduction}

The attention mechanisms, especially the transformer architectures~\cite{vaswani2017attention}, have achieved remarkable progress in Nature Language Processing (NLP) in the past few years~\cite{radford2019language, brown2020language}. Recently, the Vision Transformer (ViT) structure~\cite{dosovitskiy2020image} firstly introduces the transformers into the image classification task and suggests that a convolution-free architecture can achieve state-of-the-art performance. 
Later on, transformers have been widely used in several vision tasks performance, such as detection~\cite{carion2020end, zhu2020deformable}, segmentation~\cite{zheng2020rethinking, wang2020end}, and pose estimation~\cite{lin2020end, li2021pose}. However, the excellent performance improvement requires increasing the model size and computation complexity, and it is difficult to deploy these huge models into real-world applications like augmented reality and autonomous driving.

To address this issue, many efforts have been devoted to compressing the cumbersome transformer architectures into a lightweight counterpart, including knowledge distillation~\cite{jiao2019tinybert, sanh2019distilbert, sun2019patient, Touvron2021TrainingDI}, pruning~\cite{michel2019sixteen, chen2020lottery, chen2021chasing}, and quantization~\cite{zafrir2019q8bert, bai2020binarybert, shen2020q}. Among all these compression techniques, quantization is a popular solution as it still preserves the original network architecture. Concretely, quantization aims to replace the 32-bit parameter with a low-bitwidth representation. In NLP, the recent Binary-BERT~\cite{bai2020binarybert} pushes BERT quantization to the limit by weight binarization, and introduces quantization on activations to bring additional energy savings. 
However, compared with the one-hot tokens in language, the input image patches in vision contain richer information, and it is unclear whether the binarization is still effective in ViTs. 

In this paper, we firstly introduce the binarization into vision transformers. As fully binarization will lead to significant accuracy loss for ViTs,  we adopt binary precision for weights and low precision for activations. Different from Binary-BERT, our method directly applies the methods in 1-bit convolutions~\cite{rastegari2016xnor, liu2020reactnet} to achieve the binary weights. 
To support the inference of quantized ViT models on FPGAs, we propose a \underline{V}iT \underline{A}utomatic \underline{Q}uantization \underline{F}ramework, namely \M, to generate ViT accelerators according to the real-time frame rate requirement.
A compilation step is conducted to automatically determine the quantization precision for activations on the software side and the accelerator parameter settings on the hardware side when the model weights are compressed into binary format.
From a specific activation precision, a set of accelerator parameters can be inferred, and the overall resource utilization and inference performance can thus be estimated in advance. If the estimated frame rate meets the target, then the corresponding activation precision is decided to guide the quantization process, and the corresponding accelerator settings are adopted for hardware implementations.
Depending on the specific model structure and target frame rate, this compilation step costs several minutes to several hours, which is less than one tenth of the training time for quantization.
As for the ViT accelerator, a general compute engine is included to handle both fully-connected (FC) layers with one matrix multiplication and multi-head attention layers performing the matrix multiplication for multiple times.
Because of the binary weights, the quantized computations can be replaced with additions and subtractions, and are thus implemented with lookup tables (LUTs) on FPGAs.
The computations along the output channel, input channel, and head dimensions are totally or partially pipelined following the accelerator settings.
Besides, the data packing technique is adopted to mitigate the memory burden and improve the overall computation throughput of the implementations.

Specifically, when binarizing the weights, our quantized ViT can achieve 79.5\% on the ImageNet-1K validation set, which is just a 2.3\% performance drop compared with the full-precision model (81.8\%) but brings 32$\times$ reduction in model size. 
When further quantizing the activations to 8-bit and 6-bit, our method can still achieve accuracy of 77.6\% and 76.5\%, respectively, which outperforms previous state-of-the-art lightweight ViTs.  
The accelerator implementations are developed with Vivado High-Level Synthesis (HLS) and evaluated on the Xilinx ZCU102 FPGA board with the DeiT-base model.
The experimental results indicate that 8-bit activation quantization is necessary for an inference frame rate of 24 frames per second (FPS), and 6-bit activation quantization is needed for 30 FPS.

Our main contributions are summarized as follows: 
\begin{itemize}
    \item We firstly build the quantized vision transformer with binary weights and low precision activations and achieve new state-of-the-art performance compared with other lightweight ViTs. 
    \item We propose \M, a fully automatic framework that guides both quantization training and FPGA mapping. Given the target frame rate, \M~generates the required quantization precision and accelerator description for direct software and hardware implementations.
    \item We design a layer-specific optimization scheme, and comprehensive FPGA optimization techniques are utilized to fully explore the data efficiency, execution parallelism, and resource utilization that maximize the performance and achieve real-time execution for quantized ViT models.
\end{itemize}

The rest of the paper is organized as follows. Section~\ref{sec:related_work} discusses the related work in ViT compression and inference acceleration on hardware. Section~\ref{sec:overall_flow} provides an overall flow of \M. The ViT quantization methods are presented in Section~\ref{sec:quant_method}, and the implementation details of ViT acceleration with quantization on FPGAs are introduced in Section~\ref{sec:framework}. The experimental results are reported in Section~\ref{sec:eval}. Finally, Section~\ref{sec:conclusion} concludes the paper.

\section{Related Work} \label{sec:related_work}

\subsection{Vision Transformers}
The Vision Transformer (ViT) architecture is firstly proposed in~\cite{dosovitskiy2020image}, which uses the attention mechanism~\cite{vaswani2017attention} to solve various vision tasks. Compared to traditional CNN structures that operate on a fixed-sized window with restricted spatial interactions~\cite{raghu2021vision}, ViT allows all the positions in an image to interact through transformer blocks.
Since then, many variants have been proposed~\cite{graham2021levit,liu2021swin,yuan2021incorporating,wang2021pyramid,han2021transformer,wu2021rethinking,chen2021autoformer,steiner2021train,el2021xcit,liu2021efficient,wang2021kvt,bao2021beit}. For example, DeiT~\cite{Touvron2021TrainingDI}, T2T-ViT~\cite{yuan2021tokens} and Mixer~\cite{chen2021vision} tackle the data-inefficiency problem and make ViT trainable only with ImageNet-1K~\cite{deng2009imagenet}. PiT~\cite{heo2021rethinking} proposes a pooling-based Vision Transformer architecture with a desirable spacial dimension for better model capability and generalization performance. LV-ViT~\cite{jiang2021all} introduces a token labeling approach to improve training. PS-ViT~\cite{yue2021vision} abandons the fixed length tokens with progressive sampled tokens.

\subsection{Transformer Quantization}
Since the BERT era~\cite{vaswani2017attention}, quantization has been extensively studied to reduce the memory and computation complexity of the transformer architectures~\cite{prato2019fully, zafrir2019q8bert, shen2020q, zhang2020ternarybert, bai2020binarybert}. 
In detail,~\cite{zafrir2019q8bert} finetunes the BERT with 8-bit quantization-aware training, and successfully compresses BERT with minimal accuracy loss. To avoid severe accuracy drop,~\cite{shen2020q} uses the mixed-precision quantization. 
The later TernaryBERT~\cite{zhang2020ternarybert} proposes to use approximation based and loss-aware ternarization to ternarize the weights in the BERT, and use distillation to further reduce the accuracy drop caused by lower capacity. 
The BinaryBERT~\cite{bai2020binarybert} suggests that it is difficult to train a binary BERT directly due to its complex loss landscape, and proposes ternary weight splitting  strategy to make the binary BERT inherit the good performance of the ternary one. However, all of them are designed for NLP, not for the computer vision tasks. 
The most recent work~\cite{liu2021post} evaluates the post-training quantization on ViT and achieves comparable performance with full-precision ViT. However, they just push the compression ratio to 4$\times$ (\emph{i.e.} 8-bit) and the method is not tailored for acceleration on hardware like FPGAs.

\subsection{Lightweight ViTs}
Unlike CNNs, the ViT~\cite{dosovitskiy2020image} is typically more cumbersome and computationally extensive. 
Therefore, recently, lots of studies try to make ViT lightweight and efficient. 
These works usually revise the architecture of ViT~\cite{Touvron2021TrainingDI} and introduce the image-specific inductive bias of CNNs back into ViT~\cite{wang2021pyramid, yuan2021tokens, chen2021crossvit} to reduce the number of parameters and complexity. 
In detail, the DeiT~\cite{Touvron2021TrainingDI} introduces the distillation token and applies knowledge distillation to transfer knowledge from pre-trained networks to lightweight ViTs. 
The T2T~\cite{yuan2021tokens} proposes progressive tokenization to model the local structure information. 
The Cross-ViT~\cite{chen2021crossvit} uses a dual-branch transformer to combine image patches of different sizes to produce stronger image features, and propose a cross attention module to reduce computation. 
The MobileViT~\cite{mehta2021mobilevit} combines the strengths of CNNs and ViTs by replacing local processing in convolutions with global processing using transformers. 
All of these methods heavily modify the architecture of the standard ViT, while our quantized ViT still maintains the origin network architectures.

\subsection{Transformer Acceleration on FPGAs}
Model compression techniques has been applied and adjusted for transformer acceleration on hardware.
The approach in~\cite{li2020ftrans} leveraged block-circulant structure for weight representation and converted the matrix-vector multiplications in FC layers to FFT/IFFT computations on FPGA.
Block-based weight pruning was applied to transformers in~\cite{qi2021accommodating} with block-balanced sparsity and in~\cite{peng2021accelerating} with column balanced block-wise sparsity. The balance in these methods means that the number of sparse rows or columns in each weight block is the same, or the number of sparse blocks along each column is kept the same. This facilitates full computation resource utilization and high throughput of hardware implementations.

While pruning needs balanced sparsity for high resource usage efficiency, quantization is naturally more friendly to FPGA implementations. The method in~\cite{liu2021hardware} employed $8 \times 4$-bit and $8 \times 8$-bit quantization on different parts of BERT.
\M~differs from previous work in the following aspects: 1) The quantization process is guided by the compilation step that determines the required activation precision given the target frame rate; 2) The precision for activation quantization is chosen from a wider range to meet a specific real-time frame rate requirement.

\section{\M~Overview}
\label{sec:overall_flow}

\begin{figure*}[htb]
\centering
\includegraphics[width=0.9\textwidth]{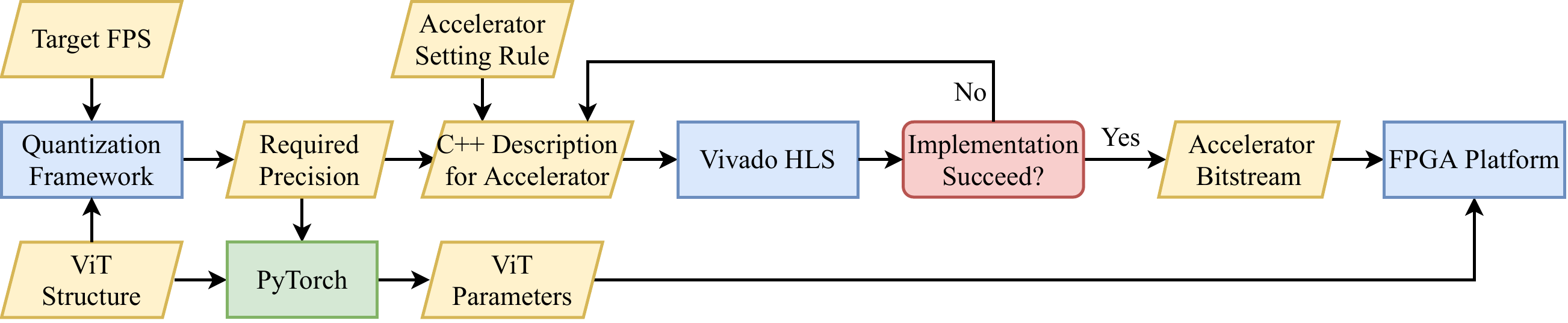}
\caption{Overall Flow of \M.}
\label{fig:framework}
\end{figure*}

Figure~\ref{fig:framework} illustrates \M~that builds an FPGA-based ViT inference accelerator. The ViT structure and desired frame rate (target FPS) are provided as input information.
A compilation step is conducted to decide the required precision for activations with the accelerator settings to satisfy the FPS target, when the weights are binary.
Specifically for an activation precision, a set of accelerator parameters can be inferred, and the overall resource utilization and inference performance can thus be estimated in advance. If the estimated frame rate meets the target, then the corresponding activation precision and accelerator settings are simultaneously decided.
On the software side, this activation precision guides the quantization with the PyTorch library, as explained in Section~\ref{sec:quant_method}, and the quantized ViT parameters are sent to the FPGA platform for model inferences.
On the hardware side, the parameter settings are adopted for the accelerator, as described in Section~\ref{sec:framework}.
This compilation step costs several minutes to several hours depending on factors such as the target frame rate, the number of model layers, and the layer dimensions. Compared with quantization that takes days for training, the compilation time is small.

The accelerator description in C++ format is then generated and synthesized with the Vivado HLS tool. The rules of setting the accelerator parameters include the initial settings to maximize the computation parallelism under the specific precision, and the adjustments of parameters if the Vivado implementation fails because of placement or routing issues. The parameters may be slightly adjusted once or twice, and a successful implementation generates a bitstream file to be deployed on FPGAs.
The synthesis and implementation for the accelerator take hours, which is also small compared with quantization training.

The key of this flow is to determine the required activation precision. For a baseline accelerator, 16-bit fixed-point numbers are used to represent the 32-bit floating-point parameters and activations of unquantized models without introducing accuracy loss.
The activation precision will therefore be chosen from range 1 to 16 bits for higher throughput than the baseline design. As the frame rate has a reciprocal relationship with the total inference time of all the model layers, maximizing the frame rate is equivalent to minimizing the model inference time. The calculation of inference time will be elaborated in Section~\ref{sec:opt_latency}.
The theoretical maximum frame rate for a ViT structure, denoted by $\mathrm{FR_{max}}$, can be obtained supposing the activation precision is 1-bit, i.e., both weights and activations are binary.
For a given target frame rate $\mathrm{FR_{tgt}}$, the feasibility for accelerator implementation is first assessed by comparing $\mathrm{FR_{tgt}}$ with $\mathrm{FR_{max}}$. $\mathrm{FR_{tgt}} \leq \mathrm{FR_{max}}$ means the accelerator supporting a frame rate no lower than $\mathrm{FR_{tgt}}$ can be implemented, and the appropriate precision is found through a binary search procedure.
With a selection range of 1 to 16 bits, up to four rounds of search are conducted to find the required precision.
In addition, if there exist multiple frame rate targets, all the possible precisions can be evaluated.

\section{ViT Quantization Method} \label{sec:quant_method}

\subsection{Preliminary: Vision Transformer} 
In this section, we revisit the architecture of the visual transformer (ViT). We give details about its each component.

\textbf{Patch Embedding} 
The ViT firstly processes a 2D image into a sequence of flattened 2D patches~\cite{dosovitskiy2020image} to adapt the 1D input sequence of the standard transformer. In detail, we denote the RGB image as $\mathbf{I} \in \mathbb{R}^{H \times W \times 3}$, where $(H, W)$ is the resolution of the original image. $\mathbf{I}$ is decomposed into $x_p \in \mathbb{R}^{N_p \times (3 \cdot P^2 ) }$, where $P$ is the resolution of each patch, and $N_p = HW / P^2$ is the total number of patches. To fit the dimension of the hidden vector $M$, ViT applies a linear transformation $W \in \mathbb{R}^{(3 \cdot P^2 ) \times M}$ to project $x_p$ into $\mathbf{X}_{I} \in \mathbb{R}^{N_p \times M}$. 
The ViT also appends the trainable \texttt{[CLS]} token $\mathbf{X}_{ \text{cls} } \in \mathbb{R}^{1 \times M}$ to represent the global feature, and adds the learnable positional embeddings $\mathbf{E}_{\text{pos}}$ to maintain positional information. Thus, the input to the transformer encoder can be represented as, 
\begin{equation}
    \mathbf{X}_0 = [ \mathbf{X}_{ \text{cls}}, \mathbf{X}_I] + \mathbf{E}_{\text{pos}}  \in \mathbb{R}^{(N_p+1) \times M}
\end{equation}
Where $[\ldots]$ is the concatenation operation. For simplicity, we denote $F = N_p + 1$ as the total number of tokens.

\begin{figure}[htb]
    \centering
    \includegraphics[width=0.99\linewidth]{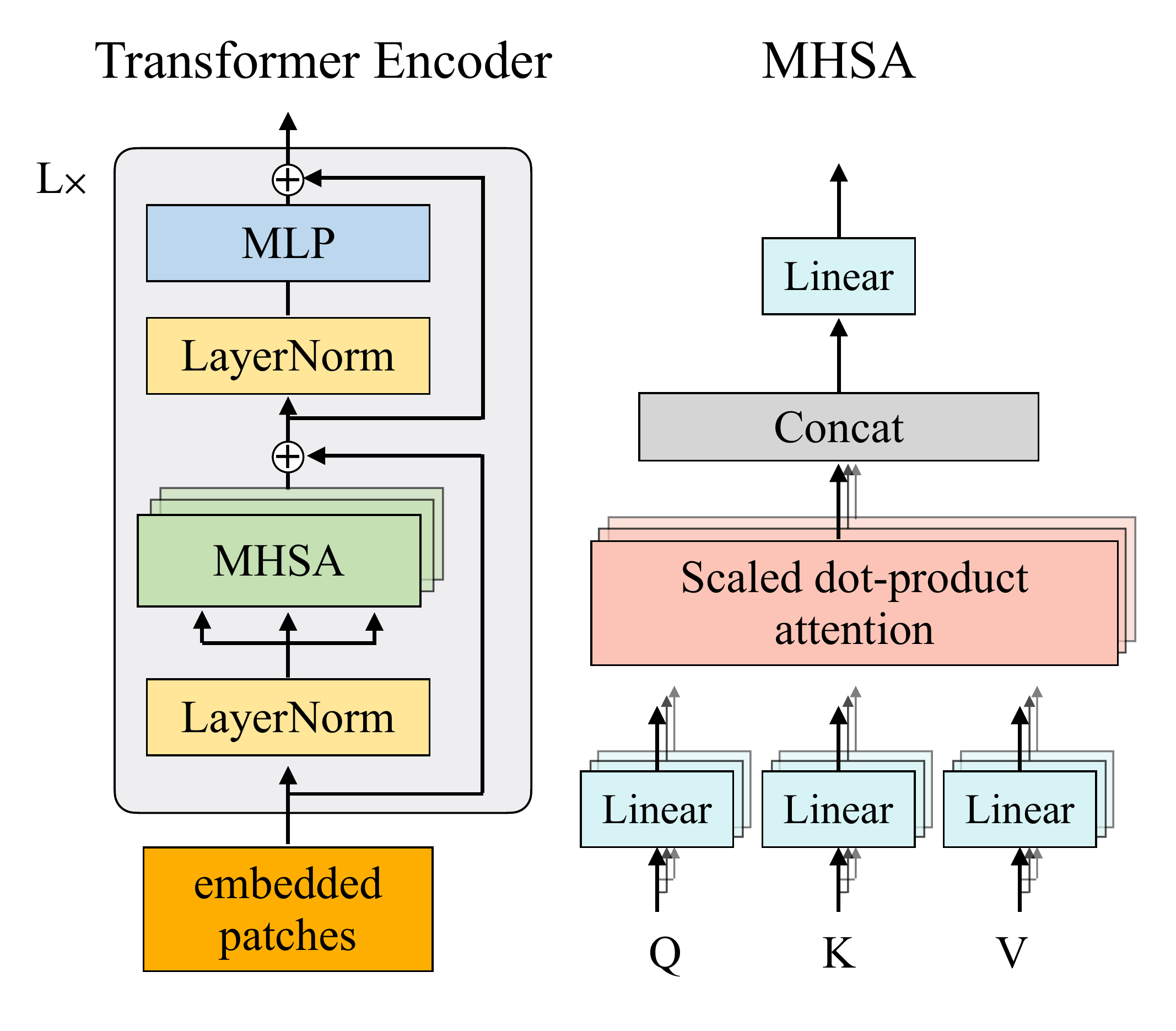}
    \caption{Illustration of transformer encoder.}
    \label{fig:encoder}
\end{figure}

\textbf{Transformer Encoder Layers}
The transformer encoder includes several layers of multi-headed self-attention (MHSA) and multi-layer perceptron (MLP) blocks. The Layernorm (LN) is applied prior to each block. Figure \ref{fig:encoder} shows the architecture of the transformer encoder layer, and operations in one encoder layer is summarized as:
\begin{equation}
\begin{aligned}
    \mathbf{X'}_l &= \text{MHSA} ( \text{LN} ( \mathbf{X}_{l-1} ) ) +  \mathbf{X}_{l-1}, & & l=1 \ldots L \\
    \mathbf{X}_l &= \text{MLP} ( \text{LN} (\mathbf{X'}_l)) + \mathbf{X'}_l, & & l=1 \ldots L
\end{aligned}
\end{equation}
In detail, the self-attention function aims to match a query and a set of key-value pairs to an output~\cite{vaswani2017attention}. Given input sequence $\mathbf{X}_l$, three linear projections $W_l^q, W_l^k, W_l^v \in \mathbb{R}^{M \times M_h}$ are firstly applied to transfer $\mathbf{X}_l$ into the query $\mathbf{Q}_l \in \mathbb{R}^{F \times M_h}$, the key $\mathbf{K}_l \in \mathbb{R}^{F \times M_h}$, and the value $\mathbf{V}_l \in \mathbb{R}^{F \times M_h}$. The attention function is calculated by
\begin{equation}
   \text{Softmax}(\mathbf{Q}_l \mathbf{K}_l^T / \sqrt{D})\mathbf{V}_l \in \mathbb{R}^{F \times M_h}
\end{equation}
Furthermore, for multi-head self-attention, $N_h$ self-attention functions are applied to the input $\mathbf{X}$, and each of them produces an output sequence. The final output is the projection of the concatenated sequences. Typically, $M_h$ is set to $M / N_h$ to maintain the consistency between the size of inputs and outputs.

The MLP block consists of two linear layers followed by a GELU activation. The first linear layer (with weights $W_l^{m1} \in \mathbb{R}^{M \times 4M}$ ) expands the dimension from $M$ to $4M$, and the second layer (with weights $W_l^{m2} \in \mathbb{R}^{4M \times M}$ ) reduces it back to $M$.

\textbf{Output Head} 
Different from the pooling layers used in computer vision, ViT directly adds the output head upon  $\mathbf{X}_L^0$, which is the representation of the \texttt{CLS} token and serves as the embedding of the entire image. The output head is defined as:  
\begin{equation}
    y = \text{LN} ( \mathbf{X}_l^0 ) W_{\text{out}},
\end{equation}
where $W_{\text{out}} \in \mathbb{R}^{M \times C}$, and $C$ is the total number of classes. 

\subsection{ViT Quantization}
Similar to~\cite{liu2020reactnet, bai2020binarybert}, we binarize the weights in ViT, and reduce the activations into low-precision, to achieve the trade-off between efficiency and accuracy.

\textbf{Binary Weights}
Following the definition in~\cite{rastegari2016xnor, liu2020reactnet}, given the matrices of real number weights $ \mathcal{W}_r$, the matrices of binary weights $\mathcal{W}_b$ are obtained by 
\begin{equation}
    w_{b} \! = \! \frac{\left\|\mathcal{W}_{r}\right\|_{l 1}}{n} \operatorname{Sign}\left(w_{r}\right)
    \! = \! \left\{ \!\! \begin{array}{l}
+\dfrac{\left\|\mathcal{W}_{r}\right\|_{l 1}}{n}, \text { if } w_{r}>0 \\
-\dfrac{\left\|\mathcal{W}_{r}\right\|_{11}}{n}, \text { if } w_{r} \leq 0
\end{array}\right.
\label{eq:binary_weight}
\end{equation}
where $w_r$ and $w_b$ denote one specific element in the matrix $\mathcal{W}_r$ and $\mathcal{W}_b$, respectively. $\frac{\left\|\mathcal{W}_{r}\right\|_{l 1}}{n}$ is the scaling factor to minimize the difference between binary and real-valued weights. 
For ViT, all trainable weights in the transformer encoder layers belong to the linear transformation. Thus, we simply replace the weights from $\mathcal{W}_r$ to $\mathcal{W}_b$ in that linear layer to obtain the activations.

\textbf{Progressive Binary Training}
As suggested by~\cite{bai2020binarybert}, training a binary BERT from scratch is challenging, as the loss landscape is steep. To alleviate this issue, we propose the progressive binary training strategy. Specifically, during the training process, we randomly select $p\%$ elements in $\mathcal{W}_r$, and only binarize these selected elements while keep other elements full-precision.   $p\%$ is set to $0 \%$ at the beginning of the training, then it grows linearly with the increase of training epoch, and achieves $100\%$ when the training process is completed. 
Therefore, $\mathcal{W}_{p}$ consists of both binary weights and real-value weights during the training process. 
Specifically, 
\begin{equation}
    \mathcal{W}_{p} = M_p \cdot \mathcal{W}_b + (1 - M_p) \cdot \mathcal{W}_r,
\end{equation}
where $M_p$ is a mask with the same size as $\mathcal{W}_r$. During training, $p\%$ elements of the mask are set to $1$, while the rest are $0$.

\textbf{Implementation Details}
Typically, both the first layer and the output head are not quantized in previous binary networks~\cite{liu2020reactnet}, as they are associated with the inputs and outputs. 
Similarly, we only consider quantizing the weights and activations within each transformer encoder, and for the patch embedding and the output head, full-precision weights and activations are utilized. 
Specifically, we binarize the weights in all attention layers ($W_l^q$, $W_l^k$, and $W_l^v$) and the weights in MLP layers ($W_l^{m1}$ and $W_l^{m2}$) with Eq. (\ref{eq:binary_weight}).  
We consider a three-step training process: 1) Train a full-precision ViT from scratch and achieve the real-valued parameters;  2) Finetune the full-precision ViT with progressive binary training to obtain ViT with binary weights and full-precision activations; 3) Finetune the binary-weight model to quantize activations with desired precision.

\section{Vision Transformer Acceleration on Hardware} \label{sec:framework}

In this section, we first discuss the optimization techniques for different types of computations in ViT layers (Sections~\ref{sec:compute_engine} and~\ref{sec:other_layers}), and then provide a detailed flow of optimizing the accelerator parameters given the model structure and desired frame rate (Section~\ref{sec:accl_framework}).

\subsection{Compute Engine for Fully-Connected and Attention Layers} \label{sec:compute_engine}

\begin{figure*}[htb]
\centering
\includegraphics[width=0.99\textwidth]{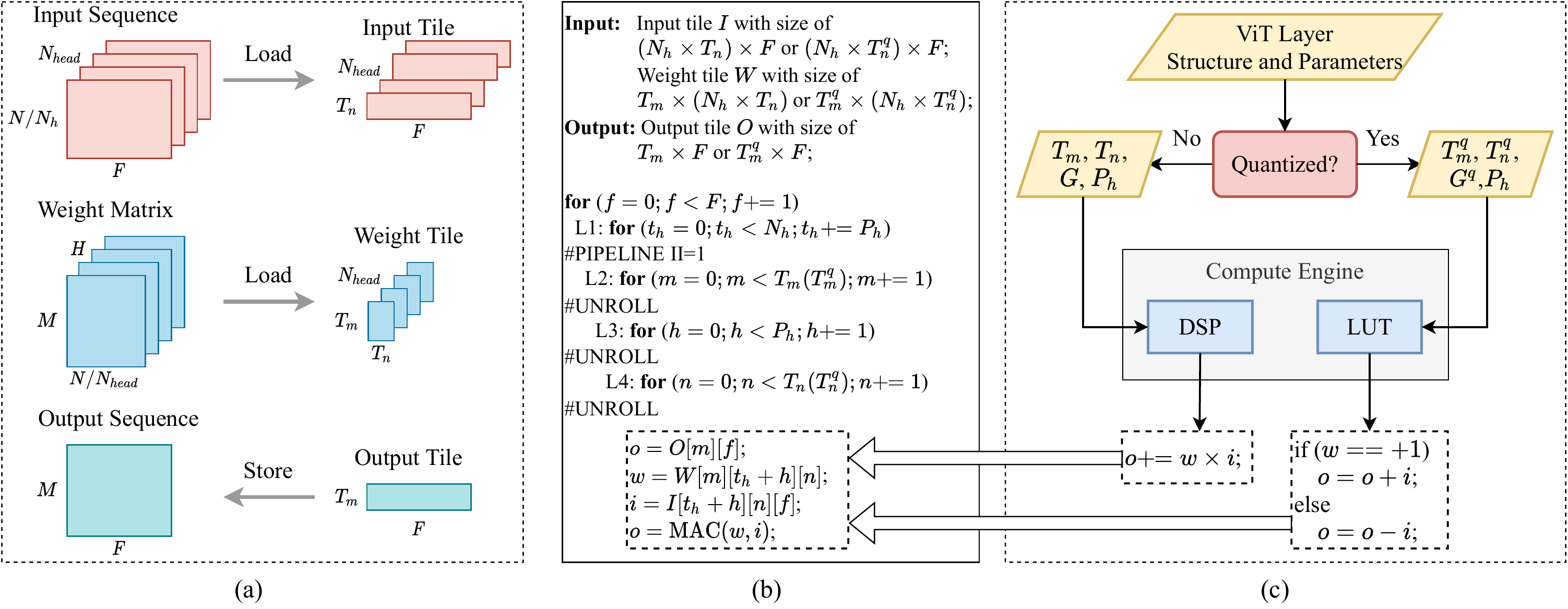}
\caption{Detailed implementation of ViT accelerator. (a) Loop tiling of input, weight, and output for one model layer; (b) Computation flow in compute engine with loop tiling, pipelining and unrolling; (c) Processing flow of one model layer based on whether it is quantized or not.}
\label{fig:compute_engine}
\end{figure*}

The implementation details of ViT acceleration on FPGA are displayed in Figure~\ref{fig:compute_engine}, and the notations of variables and parameters are listed in Table~\ref{tab:notation}.
As the storage and computation resources are limited on FPGA devices, the loop tiling technique is adopted to split the input, weight, and output data for each ViT layer into tiles, and tiling for an unquantized layer is shown as an example in Figure~\ref{fig:compute_engine}(a).

\begin{table}[htb]
\centering
\small

\caption{Notations for the ViT accelerator.}
\label{tab:notation}

\begin{tabular}{l|l}
\toprule
Notation & Description \\
\midrule
$M$ ($N$) & Number of output (input) channels \\ \hline
$F$ & Number of token sequences \\ \hline

\multirow{2}{*}{$T_m$ ($T_m^q$)} & Tiling size for unquantized (quantized) data \\
~ & in output channel dimension \\ \hline
\multirow{2}{*}{$T_n$ ($T_n^q$)} & Tiling size for unquantized (quantized) data \\
~ & in input channel dimension \\ \hline
$N_h$ & Total number of heads \\ \hline
$P_h$ & Number of heads for computation in parallel \\ \hline
\multirow{2}{*}{$G$ ($G^q$)} & Number of unquantized (quantized) data \\
~ & packed as one \\ \hline
$S_{port}$ & Size of each AXI port on FPGA \\ \hline

$p_{in}$ & Number of AXI ports used for data transfer \\
($p_{out}$, $p_{wgt}$) &  of input (output, weight) tile \\ \hline
$J_{in}$& Number of clock cycles for input transfer \\
($J_{wgt}$, $J_{out}$, & (weight transfer, output transfer, \\
$J_{cmpt}$) & computation) for a group of tiles \\ \hline
$B_{in}$ & Number of BRAMs used by input (output, \\
($B_{out}$, $B_{wgt}$) & weight) tile \\ \hline
$S_{bram}$ & Available number of BRAMs (DSPs, LUTs) \\
($S_{dsp}$, $S_{lut}$) & on FPGA \\

\bottomrule
\end{tabular}
\end{table}

The most computation-intensive layers in ViTs include FC layers that exist in both multi-layer perceptron (MLP) modules and multi-head attention modules, and scaled dot-product attention layers that appear in multi-head attention modules. The primary computations of these two types of layers are both matrix multiplications.
While an FC layer performs only one matrix multiplication, multi-head attention repeats the computations multiple times in parallel, so the accelerator is designed such that the computations can be executed in parallel across $P_h$ attention heads.
To make the design also compatible for FC computations, the $N$ input channels in an FC layer are split into $N_h$ groups, $P_h$ of which are processed by the compute engine simultaneously at a time.
A control signal is added to indicate whether the current layer is multi-head attention. For multi-head attention, the $N_h$ heads of results are kept as they are, while for an FC layer, these results from $N_h$ input channel groups are added together, making the final result for each output channel accumulated from all the input channels.

The main computation flow general for these two types of layers is displayed in Figure~\ref{fig:compute_engine}(b). The loops L2, L3, L4 under L1 are unrolled and pipelined so that the compute engine can manage $T_m \cdot P_h \cdot T_n$ multiply-accumulate (MAC) operations in parallel.
For illustration simplicity of the MAC operation with the input and weight, the head and the input channel dimensions are shown separately.
Figure~\ref{fig:compute_engine}(c) provides the processing flow of one ViT layer with or without quantization.
The accelerator parameters needed for the compute engine include $T_m$ ($T_m^q$), $T_n$ ($T_n^q$), $G$ ($G^q$), and $P_h$, the settings and adjustments of which are elaborated in Section~\ref{sec:accl_framework}. 
Unquantized computations with high precision are managed by the DSP resources on FPGA, whereas quantized computations can be replaced with additions or subtractions that would be managed by LUTs, since the weight value is binarized as either $+1$ or $-1$.

\subsection{Processing of Other Layers in Vision Transformer} \label{sec:other_layers}

\begin{figure}[tb]
\centering
\includegraphics[width=0.8\columnwidth]{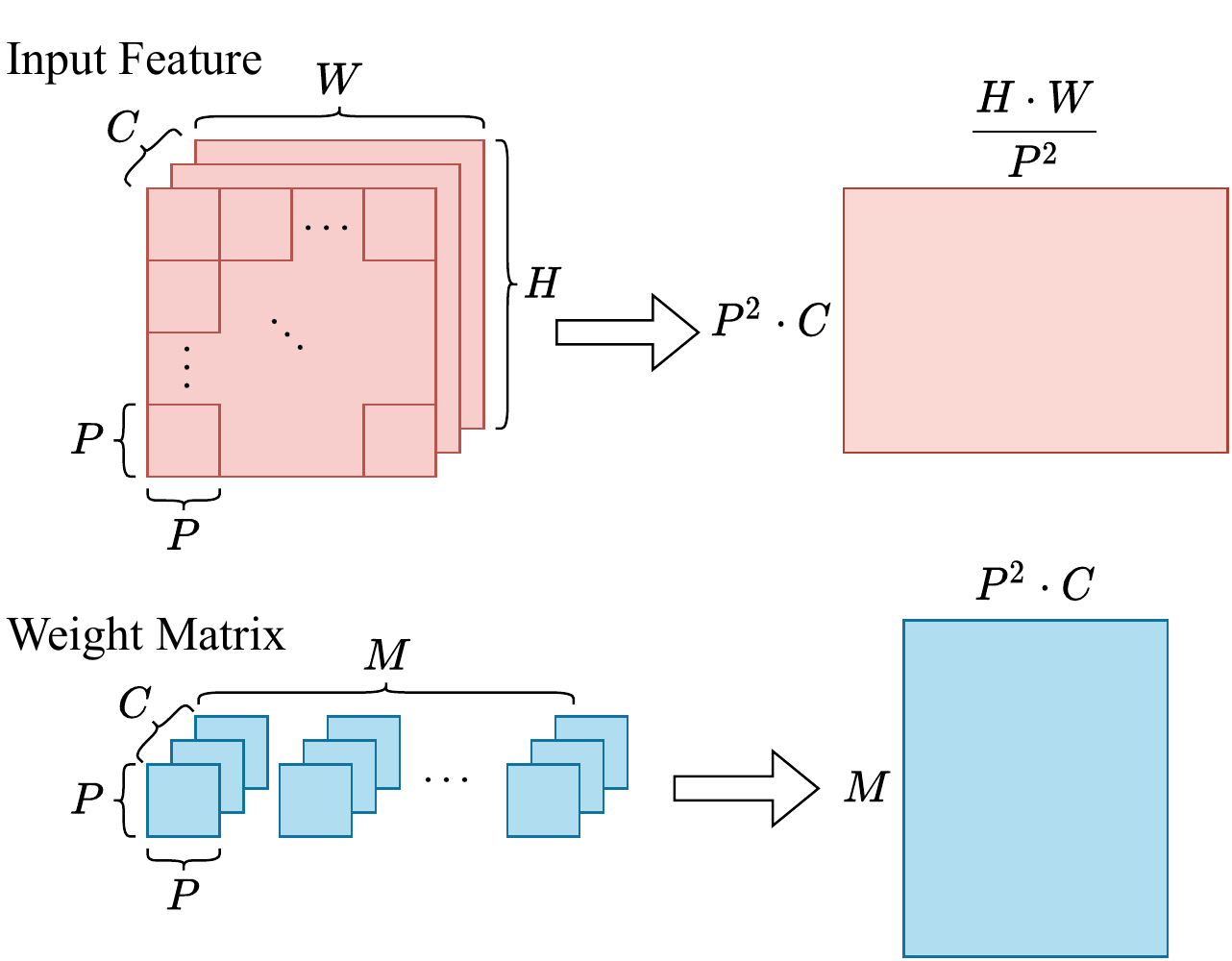}
\caption{Conversion of input and weight tensors for the first convolutional layer.}
\label{fig:patch_embed}
\end{figure}

In addition to matrix multiplications in MLP and multi-head attention modules, ViTs contain convolution, scaling, softmax, activation, normalization, and skip-connection addition operations.
The first layer of a ViT is a convolutional layer that can be converted to an FC layer, because its kernel size and stride is the same as the patch size $P$, meaning that the input data are used only once when a weight kernel slides across the input feature map,
as shown in Figure~\ref{fig:patch_embed}, where $C$, $H$ and $W$ denote the number of input channels, the height and width for the original input feature.
The scaling, softmax, and GELU activation operations are performed on the host CPU of the FPGA, which introduces very small latency overhead for embedded FPGAs compared with matrix multiplications.

\subsubsection{Processing of Normalization Layers and Skip-Connections}

In ViTs, layer normalization is applied at the beginning of each multi-head attention and each MLP module, and there is an identity skip-connection linking the input activations of each normalization layer and the output activations of the subsequent module, which can be seen from Figure~\ref{fig:encoder}.
In the sequential processing of the layers, the inputs of the normalization layer require to be stored for later additions with the outputs of the following multi-head attention or MLP module.
Since normalization layers are not so compute-intensive as FC ones, their parameters and inputs are not quantized but rather kept with higher precision, namely 16-bit on hardware to preserve the model accuracy.
As a result, two data transfer ports are needed, one for the inputs of normalization that are stored as unquantized data, and the other for the outputs of normalization, also the inputs for the subsequent FC layer, which are stored as quantized data.
It needs to be emphasized that the data transfer port for quantized FC inputs is necessary to minimize the input loading time for the FC layer as the input tiles would be loaded multiple times in the matrix multiplication with loop tiling.

\subsection{Design of FPGA-Based Vision Transformer Accelerator} \label{sec:accl_framework}

Before building the ViT accelerator supporting a specific frame rate, a baseline accelerator is realized for unquantized models, whose 32-bit floating-point parameters and activations are represented with 16-bit fixed-point numbers to reduce the computation and storage resource utilization without accuracy loss on hardware. Let us denote the optimized parameters for the baseline design as $T_m^{base}$, $T_n^{base}$ and $G^{base}$.
These parameters are treated as the starting point to find the required activation precision and optimize the ViT accelerator parameters.

Provided the target frame rate, \M~will determine the proper activation precision, which is an inverse procedure, as the activation precision directly affects the accelerator settings and hence the actual frame rate.
The following discussion of hardware implementation details is on the basis of a specific activation precision, but it does not mean the proper precision has been decided in the first step. Instead, it is found among several possible precisions after the analysis in terms of resource utilization and inference performance.

\subsubsection{Data Packing} \label{sec:datapack}

The data packing technique is employed to reduce the block RAM (BRAM) usage and the latency of data transfer between off-chip memory (DDR) and on-chip memory (BRAM).
Although each BRAM on Xilinx FPGAs can accommodate 18k-bit data, the space of the whole BRAM might not be fully utilized, leading to high BRAM usage in the ViT acceleration implementation. This mainly results from loop pipelining and unrolling that requires each related data array to be partitioned into multiple smaller arrays.
Data packing can mitigate this issue through concatenating multiple low-precision numbers as one number. With a data packing factor of $G$, the overall BRAM usage can be reduced by up to $G$ times, and the number of clock cycles for input loading and output storage can be reduced by $G$ times as well.

Each AXI port on FPGA with size $S_{port}$ can accommodate $G$ unquantized data or $G^q$ quantized data. If $S_{port}=64$, then $G=4$ for 16-bit data used in our baseline design, and $G^q=8$ if the quantization precision is 8-bit. A special case is when $S_{port}$ cannot be divided exactly by the bit-length of quantized values. Take 6-bit quantization precision as an instance, $G^q = \left\lceil \frac{64}{6} \right\rceil = 10$, and only 60 of the 64 bits are exploited.
Data packing is performed in the input channel dimension of the weights and input activations, and in the output channel dimension of the output activations. It is not displayed in Section~\ref{fig:compute_engine}(b) for illustration simplicity of MAC operation.

\subsubsection{Determing Parameters for Best Computation Parallelism} \label{sec:accl_param}

Although most layers of the ViT model are quantized, the first and last layers are not quantized to better maintain the model accuracy, and the accelerator compute engine still needs to handle unquantized data. As a result, we have two groups of accelerator parameters to be determined, i.e., $T_m$, $T_n$ and $G$ for unquantized layers, and $T_m^q$, $T_n^q$ and $G^q$ for quantized ones, while the parameter $P_h$ remains the same in both situations.
The ViT layers are handled one by one, so the accelerator will not perform unquantized computations and quantized ones simultaneously. Nonetheless, the same BRAMs for input, weight and output data can be utilized whether the layer is quantized or not. The related accelerator parameters are therefore decided to make the best effort to utilize the BRAMs.

When creating the ViT accelerator with support of quantization, the accelerator parameters for unquantized layers are first set as $T_n = T_n^{base}$ and $G = G^{base}$, and $T_m$ is initially set to a value that is near to $T_m^{base}$ and can be divided exactly by $G$ and $G^q$.
As mentioned in Section~\ref{sec:datapack}, $G^q$ is directly calculated according to $S_{port}$ and the required quantization precision. $T_n^q$ is then calculated by $T_n^q = \left\lfloor T_n \cdot \frac{G_q}{G} \right\rfloor$ for maximum utilization ratio of BRAMs for quantized data. $P_h$ is usually a value that can divide $N_h$ exactly. For instance, if $N_h=6$, $P_h$ is set to 3; if $N_h=8$ or $N_h=12$, then $P_h$ is 4. In the next step, $T_m^q$ is set as equal to $T_m$, and a hardware design with all these parameters is synthesized and implemented with Vivado HLS as an initial try. If the FPGA board cannot accommodate this design due to placement or routing issues (usually resulting from overutilization of LUTs), then $T_m$ and $T_m^q$ are adjusted for the implementation to fit into the board and minimize the overall inference latency as well. For low activation precision, $T_m$ is reduced and $T_m^q$ is increased until the FPGA resources are fully exploited. In this procedure, both $T_m$ and $T_m^q$ are kept as values that can be divided exactly by $G$ and $G^q$ for convenience of output storage.

\subsubsection{Performance Analysis and Objective Function} \label{sec:opt_latency}

The main variables in the equations in this section are explained in Table~\ref{tab:notation}.
For one layer $i$ in ViT, the numbers of clock cycles needed for input tile loading, weight tile loading, and output tile storage are calculated as
\begin{equation}
\small
\begin{aligned}
J_{in} &= N_h \cdot \left( (1-\alpha) \cdot \left\lceil \frac{T_n}{G} \right\rceil + \alpha \cdot \left\lceil \frac{T_n^q}{G^q} \right\rceil \right) \cdot \left\lceil \frac{F}{p_{in}} \right\rceil, \\
J_{wgt} &= N_h \cdot \left( (1-\alpha) \cdot \left\lceil \frac{T_n}{G} \right\rceil + \alpha \cdot \left\lceil \frac{T_n^q}{G^q} \right\rceil \right) \cdot \left\lceil \frac{T_m}{p_{wgt}} \right\rceil, \\
J_{out} &= (1 + \gamma) \! \cdot \! \left( (1-\beta) \! \cdot \! \left\lceil \frac{T_m}{G} \right\rceil \! + \! \beta \! \cdot \! \left\lceil \frac{T_m^q}{G^q} \right\rceil \right) \! \cdot \! \left\lceil \frac{F}{p_{out}} \right\rceil,
\end{aligned}
\end{equation}
where $\alpha$ is 1 if the inputs and weights are quantized else 0, $\beta$ is 1 if the outputs are quantized else 0, and $\gamma$ is $N_h-1$ if the current layer is a multi-head attention layer else 0.
Additionally, the clock cycle number of computations for one group of tiles is
\begin{equation}
\small
J_{cmpt} = F \cdot \left\lceil \frac{N_h}{P_h} \right\rceil.
\end{equation}
The data loading and computation for the tiles are conducted simultaneously with the double buffering technique to overlap the data transfer with computations. The clock cycle number of this process is
\begin{equation}
\small
J_{lc} = \max\{J_{in}, J_{wgt}, J_{cmpt}\}.
\end{equation}
And to obtain the accumulation of output results, this process is performed multiple times. The clock cycle number for calculating the whole output tile is
\begin{equation}
\small
\begin{aligned}
J_s =& \max \bigg\{J_{lc} \cdot \bigg( (1-\alpha) \cdot \left\lceil \frac{N}{N_h \cdot T_n} \right\rceil \\
& + \alpha \cdot \left\lceil \frac{N}{N_h \cdot T_n^q} \right\rceil \bigg) + J_{cmpt}, J_{out} \bigg\}.
\end{aligned}
\end{equation}
The overall clock cycle number for a ViT layer $i$ is therefore described by
\begin{equation}
\small
J_i = \left( (1-\beta) \cdot \left\lceil \frac{M}{T_m} \right\rceil + \beta \cdot \left\lceil \frac{M}{T_m^q} \right\rceil \right) \cdot J_s + J_{out}.
\end{equation}

With double buffering, the 18k-bit BRAM usage of the input, weight, and output tiles are given by
\begin{equation}
\scriptsize
\begin{aligned}
B_{in} &= 2 \cdot N_h \cdot \max \! \left\{ \! \left\lceil \frac{T_n}{G} \right\rceil \! \cdot \! \left\lceil \frac{F \cdot G \cdot 16}{18\mathrm{k}} \right\rceil, \left\lceil \frac{T_n^q}{G^q} \right\rceil \! \cdot \! \left\lceil \frac{F \cdot G^q \cdot b^q}{18\mathrm{k}} \right\rceil \! \right\}, \\
B_{wgt} &= 2 \cdot N_h \cdot \max \! \left\{ \! \left\lceil \frac{T_n}{G} \right\rceil \! \cdot \! \left\lceil \frac{T_m \! \cdot \! G \cdot 16}{18\mathrm{k}} \right\rceil, \left\lceil \frac{T_n^q}{G^q} \right\rceil \! \cdot \! \left\lceil \frac{T_m \! \cdot \! G^q}{18\mathrm{k}} \right\rceil \! \right\}, \\
B_{out} &= 2 \cdot N_h \cdot \max \! \left\{ \! \left\lceil \frac{T_m}{G} \right\rceil \! \cdot \! \left\lceil \frac{F \cdot G \cdot 16}{18\mathrm{k}} \right\rceil, \left\lceil \frac{T_m^q}{G^q} \right\rceil \! \cdot \! \left\lceil \frac{F \cdot G^q \cdot b^q}{18\mathrm{k}} \right\rceil \! \right\},
\end{aligned}
\end{equation}
where $b^q$ is the activation bit-width in quantization. As for the DSP utilization, since each DSP can manage one unquantized MAC operation with a 16-bit input and a 16-bit weight, the number of used DSPs is calculated by $T_m \cdot P_h \cdot T_n$ to perform $T_m \cdot P_h \cdot T_n$ MAC operations in parallel.

Finally, our objective function can be written as
\begin{equation}
\textrm{minimize} \quad \sum\limits_i J_i
\end{equation}
subject to
\begin{equation}
\begin{aligned}
B_{in} + B_{wgt} + B_{out} &\leq S_{bram}, \\
T_m \cdot P_h \cdot T_n &\leq S_{dsp} \cdot r_{dsp}, \\
C_{lut} \cdot T_m^q \cdot P_h \cdot T_n^q &\leq S_{lut} \cdot r_{lut},
\end{aligned}
\end{equation}
where $r_{dsp}$ and $r_{lut}$ are respectively the maximum ratio of DSPs and LUTs to be utilized for MAC operations, and $C_{lut}$ is the LUT cost for one MAC with quantized operands.
As mentioned in Section~\ref{sec:overall_flow}, minimizing the overall model latency is equivalent to maximizing the frame rate. If the maximum frame rate is no lower than the target frame rate, then the accelerator with the corresponding activation precision and parameter settings is feasible.

\section{Evaluation} \label{sec:eval}

\subsection{Experimental Setup}

\textbf{Datasets \& Architectures}
We conduct all experiments on ImageNet-1K~\cite{deng2009imagenet} and use the top-1 accuracy on validation set as the evaluation metric. All images are resized to $224$. 
We use the DeiT-base without the distillation token as the default ViT architecture~\cite{Touvron2021TrainingDI}.
We consider three types of quantization precision: 1) binary weights, and full-precision activations (W1A32); 2) binary weights, and 8-bit activations (W1A8); 3) binary weights, and 6-bit activations (W1A6).

\textbf{Training}
We strictly follow the training settings in~\cite{Touvron2021TrainingDI} for all the three training stages. In detail, the network is trained for $300$ epochs and is optimized by AdamW~\cite{loshchilov2018decoupled} with weight decay $0.05$. The batch size is set to $512$, The learning rate is set to $5 \times 10^{-4}$ initially and is decayed with a cosine annealing schedule. All data augmentations in~\cite{Touvron2021TrainingDI} are also included during the training process.

\textbf{Hardware}
The ViT accelerators with different precisions are implemented on the Xilinx ZCU102 FPGA platform with 2520 DSPs and 274k LUTs, while \M~can be generalized to other types of FPGAs. 
For all the implementations, the operating frequency is set to 150~MHz to maximize the computation efficiency and avoid timing violation.
The hardware implementations are preformed with the HLS tool of Xilinx Vivado 2020.1.

\subsection{Results on Software}
\subsubsection{Comparison with Other Lightweight ViTs} 
We first compare our quantized DeiT-base with other lightweight ViTs that are trained from scratch on the ImageNet-1K dataset. All the models are trained without additional distillation. 
We also report the space usage of devices for each model/method, which is calculated as the product of the number of parameters and the precision. 
All results are shown in Table \ref{tab:results}. 
Firstly, all other lightweight ViTs are still full-precision models, while the weights in our method are all binarized. It can be seen that with binarized weight, it is feasible to consume much less storage space of devices without changing the original network architecture.  
Secondly, compared with the full-precision DeiT-base,  our weight-binarized version only drops 2.3\% accuracy (from 81.8\% to 79.5 \%). When further quantizing the activations to low percision, the accuracy can still maintain 77.6\% for 8-bit and and 76.5\% for 6-bit, which is still much better than other lightweight ViTs. Although quantizing the activations does not reduce the model size on devices, it can reduce the inference time on the hardware. More details of the efficiency evaluation will be discussed in Section \ref{sec: results_hardware}. 

\begin{table}[htb]
\tabcolsep 3pt
    \centering
    \small
    \caption{Comparison of ViT variants on ImageNet validation set.}
    \resizebox{0.99\linewidth}{!}{
     \begin{tabular}{l|c|c}
    \toprule
        Method           & Accuracy (\%) &  Space Usage \\
        \midrule
        DeiT-base                              & 81.8   & 86M $\times$ 32 \\
        \midrule
        T2T~\cite{yuan2021tokens}           & 71.7   & 4.7M $\times$ 32 \\
        DeiT~\cite{Touvron2021TrainingDI}    &  72.2   & 5.7M $\times$ 32  \\
        PiT~\cite{heo2021rethinking}        & 73.0   & 4.9M $\times$ 32  \\
        Cross-ViT~\cite{chen2021crossvit}   &  73.4  & 6.9M $\times$ 32  \\
        MobileViT ~\cite{mehta2021mobilevit} & 74.8 & 2.3M $\times$ 32  \\
        \midrule
        Ours (DeiT-base-W1A32) &   \textbf{79.5}      &  86M $\times$ 1 \\
        Ours (DeiT-base-W1A8) &  77.6 &  86M $\times$ 1  \\
        Ours (DeiT-base-W1A6) &  76.5 &  86M $\times$ 1  \\
    \bottomrule
    \end{tabular}   
    }
    \label{tab:results}
\end{table}

\subsubsection{Ablation Studies}

\textbf{Architecture of ViT} 
Besides DeiT-base, we also evaluate the accuracy of mix-quantization on DeiT-tiny and DeiT-small~\cite{Touvron2021TrainingDI} on ImageNet. In detail, the number of parameters of DeiT-tiny is $5$ million, and that of DeiT-small is $22$ million, due to the smaller number of embedding dimension and fewer heads. 
Table \ref{tab:ablation_archs} presents the comparison between the full-precision model (W32A32) and the model with binary weights and full-precision activations (W1A32). Different from the DeiT-base, after binarizing their weights, the accuracy of both DeiT-tiny and DeiT-small drop heavily. 
We hypothesize that these lightweight networks are very fragile as the number of parameters is quite limited. As a consequence, when binarizing their weights, the learning capacity is heavily restricted and thus it is challenging to learn the spatial inductive bias for vision from scratch~\cite{dosovitskiy2020image}. 
Therefore, it is difficult to binarize these lightweight ViTs and maintain their accuracy. 

\begin{table}[htb]
    \centering
    \small 
    \caption{Accuracy (\%) of mix-quantization DeiT-tiny and DeiT-small on ImageNet validation set.}
   
    \begin{tabular}{l|cc}
    \toprule
    Quantization Precision          & DeiT-tiny & DeiT-small \\
    \midrule
    W32A32           & 72.2     & 79.9      \\
    W1A32 & 51.5     & 70.4     \\
    \bottomrule
    \end{tabular}
    \label{tab:ablation_archs}
\end{table}

\textbf{Training Schedules}
In this section, we aim to show the effect of three-stage training steps and progressive binarization strategy. Due to the computing limitation, we randomly sample 100 categories from the full ImageNet datasets (named ``ImageNet-100"), and conduct the ablation studies on this subset. Specifically, we take 32-bit pre-training or progressive binary training out of the training procedure to evaluate the effectiveness of each strategy. Table \ref{tab:ablation_training} shows the ablation results. All the results are based on DeiT-small architecture. 
Remarkably, without the full-precision pre-training, there is a huge drop of accuracy. When further removing the progressive binarization, there is an additional 0.9\% accuracy drop. These results verify the effectiveness of our proposed stage-wise finetuning strategy and progressive binarization.

\begin{table}[htb]
    \centering
    \small 
    \caption{Ablation studies on ImageNet-100 validation set.}
  
    \begin{tabular}{l|c}
    \toprule
    Method   & Accuracy (\%) \\
    \midrule
    W1A32 & 84.3 \\
    W1A32 (w/o  pre-training)  &   79.3      \\
    W1A32 (w/o progressive ) & 78.4     \\
    \bottomrule
    \end{tabular}
    \label{tab:ablation_training}
\end{table}

\subsection{Results on Hardware} \label{sec: results_hardware}

\begin{table*}[htb]
\centering
\small

\caption{Hardware resource utilization and performance of ViT accelerators with different frame rates and precisions.}
\label{tab:util_perf}

\begin{tabular}{c|cccc|cccc}
\toprule
\multirow{2}{*}{\makecell{Quantization \\ Precision}} &
\multicolumn{4}{c|}{Resource Utilization} & \multirow{2}{*}{FPS} & \multirow{2}{*}{\makecell{Throughput \\ (GOPS)}} & \multirow{2}{*}{GOPS/DSP} & \multirow{2}{*}{GOPS/kLUT} \\
\cline{2-5}
~ & DSP & kLUT & BRAM36 & kFF & \\
\midrule
W32A32 & 1564 (62\%) & 120 (44\%) & 453.5 (50\%) & 99 (18\%) & 10.0 & 345.8 & 0.221 & 2.882 \\
W1A8 & 1564 (62\%) & 143 (52\%) & 565.5 (62\%) & 110 (20\%) & 24.8 & 861.2 & 0.551 & 6.022 \\
W1A6 & 673 (27\%) & 166 (60\%) & 392.5 (43\%) & 82 (15\%) & 31.6 & 1096 & 1.628 & 6.599 \\
\bottomrule
\end{tabular}

\end{table*}

\begin{table*}[htb]
\tabcolsep 4pt
\centering
\small

\caption{Performance comparison among FPGA accelerators, CPU and GPU.}
\label{tab:compare_cpu_gpu}

\begin{tabular}{c|c|c|cc|ccc}
\toprule
\multirow{2}{*}{\diagbox{Performance}{Implementation}} & CPU & GPU & \multicolumn{2}{c|}{\cite{liu2021hardware} for BERT} & \multicolumn{3}{c}{Ours for DeiT-base} \\
& i7-9800X & TITAN RTX & ZCU102 & ZCU111 & W32A32 & W1A8 & W1A6 \\
\midrule
FPS & 15.3 & 183.4 & 22.8 & 42.0 & 10.0 & 24.8 & 31.6 \\
Power (W) & 100 & 260 & 9.8 & 13.2 & 9.9 & 8.7 & 7.8 \\
Energy Efficiency (FPS/W) & 0.15 & 0.71 & 2.32 & 3.18 & 1.01 & 2.85 & 4.05 \\
\bottomrule
\end{tabular}
\end{table*}

\subsubsection{Resource Utilization and Performance of ViT Accelerators}

Table~\ref{tab:util_perf} summarizes the resource utilization and performance of the \M-generated ViT accelerators with various frame rates and precisions on FPGAs.
The quantization precision is represented by W[$q_w$]A[$q_a$], where $q_w$ is the bit-length for weights and $q_a$ the bit-length for activations.
For quantized models, the precision on hardware is the same as that on software, while for unquantized models, the parameters and activations in W32A32 precision on software are represented with W16A16 precision on hardware to reduce the computation and storage resource utilization without accuracy loss.
The utilization of resources on FPGA is shown with the percentage of the used and total available number. The numbers of LUTs and FFs are expressed as thousands, and the number of BRAMs is listed treating the capacity of each BRAM as 36k bits.
The performance of ViT implementations are obtained from the inference process of DeiT-base in terms of the frame rate in FPS, the throughput in giga operations per second (GOPS), and computation efficiency including GOPS per DSP and GOPS per thousands of LUTs.

The inference speed of the W1A8 design is 24.8 FPS, and that for W1A6 is 31.6 FPS.
Compared with the baseline implementation with W32A32, these two designs respectively achieve 2.48$\times$ and 3.16$\times$ acceleration.
In these low-precision accelerators, the parallelism in unquantized computations is degraded because of the reduction of DSP utilization, while that in quantized computations is enhanced since more LUTs are utilized.
This assists the accelerators to attain higher overall throughput and better fit into the available computation resources, avoiding placement and routing issues.
Higher throughput and lower DSP utilization results in 2.49$\times$ higher GOPS/DSP of the W1A8 design than the W32A32 one, and this ratio for W1A6 can reach 7.37$\times$.
With slightly more LUT utilization for the quantized operations and extra logic to select between unquantized or quantized operations, the GOPS/kLUT increases by 2.09$\times$ and 2.29$\times$ for W1A8 and W1A6 designs, respectively.

These experimental results demonstrate that the accelerators generated by \M~are able to achieve real-time inferences, i.e., a frame rate requirement of 24 FPS is satisfied with 8-bit quantization for activations, and a target of 30 FPS is met with 6-bit activation quantization.
\M~can be further generalized to other frame rate targets and other types of transformers.

\subsubsection{Comparison with Other Implementations}

Our accelerators are further compared with previous work, CPU and GPU with regard to FPS, power, and energy efficiency. Since no study using quantization has been carried out for ViT acceleration on FPGAs, the accelerators for BERT in~\cite{liu2021hardware} with $8 \times 4$-bit and $8 \times 8$-bit quantization are used for comparison, and other implementation results are all obtained for DeiT-base. As shown in Table~\ref{tab:compare_cpu_gpu}, our W1A8 and W1A6 accelerators both outperform the BERT design on ZCU102 in terms of FPS, power and energy efficiency, and the W1A6 design has the highest FPS/W among all implementations. Compared with the inferences on Intel(R) Core(TM) i7-9800X CPU and NVIDIA TITAN RTX GPU, our W1A6 design has 27.0$\times$ and 5.7$\times$ improvement in FPS/W.

\section{Conclusion} \label{sec:conclusion}

This work proposes a framework called \M~for automatic construction of inference accelerators for ViTs with binary weights and low-precision activations under the desired frame rate. 
\M~first performs compilation to decide the required activation precision with the accelerator settings related to the computation parallelism by analyzing the resource utilization and inference performance.
And then for the accelerator implementation, a compute engine general for FC layers and multi-head attention layers is designed, and the quantized computations are replaced with additions and subtractions that are implemented with LUTs.
Optimization techniques like data packing and loop pipelining are utilized with different factors for unquantized and quantized computations to enhance the computation parallelism and memory utilization efficiency.
The experimental results for the DeiT-base model show that the accelerator with 8-bit activation quantization achieves an inference speed of 24.8 FPS, and that with 6-bit activation quantization attains 31.6 FPS, which can meet various real-time requirements.

\bibliography{main}
\bibliographystyle{abbrv}


\end{document}